\pdfoutput=1

\documentclass[11pt]{article}

\usepackage[final]{acl}

\usepackage{times}
\usepackage{latexsym}
\usepackage{graphicx}
\usepackage[T1]{fontenc}

\usepackage[utf8]{inputenc}

\usepackage{microtype}
\usepackage{natbib}

\title{Towards Compositionally Generalizable Semantic Parsing\\ in Large Language Models: A Survey}

\author{Amogh Mannekote \\ University of Florida} 

\begin{document}

\maketitle

\begin{abstract}
    Compositional generalization is the ability of a model to generalize to complex, previously unseen types of combinations of entities from just having seen the primitives.
    This type of generalization is particularly relevant to the semantic parsing community for applications such as task-oriented dialogue, text-to-SQL parsing, and information retrieval, as they can harbor infinite complexity.
    Despite the success of large language models (LLMs) in a wide range of NLP tasks, unlocking perfect compositional generalization still remains one of the few last unsolved frontiers. 
    The past few years has seen a surge of interest in works that explore the limitations of, methods to improve, and evaluation metrics for compositional generalization capabilities of LLMs for semantic parsing tasks.
    In this work, we present a literature survey geared at synthesizing recent advances in analysis, methods, and evaluation schemes to offer a starting point for both practitioners and researchers in this area.
\end{abstract}

\section{Introduction}

    Semantic parsing is a general-purpose task that involves translating natural language input into a structured (and usually symbolic) output. The real-world applications of semantic parsing are wide-ranging and often mission-critical. They range from task-oriented dialogue to text-to-SQL business applications to information retrieval. Compositionality is the ability of a model to understand previously unseen combinations ``jump left'' (compound) of primitive concepts (atoms) such as ``jump`` and ``left'' without ever having been trained on the compound directly.

    Today, large language model based semantic parsers dominate the semantic parsing scene in both research and applied settings owing to the unprecedented representational power and flexibility that they offer for capturing rich natural language as compared to their predecessors.
    However, despite demonstrating state-of-the-art results over standard benchmarks, careful error analyses show that these models fail to generalize to cases where the output is a symbolic form containing nested structures of depths greater than what is seen by the models during training \cite{3-yao-structural-generalization}.
    
    More generally, although existing benchmark corpora aid in evaluating the strengths and weaknesses of a model, they only offer a coarse-grained view of the its underlying capabilities.
    To overcome this coarseness, the semantic parsing community has expended significant research effort to develop corpora with more purposeful train-test splits, with the aim of leveraging them to run more fine-grained experiments.

The task of categorizing works that seek to quantify and improve the compositional generalization capabilities of LLMs is quite challenging. Nevertheless, we begin to ``chip away at the block'' by looking at some of the key guiding philosophies adopted by contemporary works.

Several lines of research stem from the semantic parsing community's dissatisfaction with the performance of ``vanilla training'' of Transformer-based LLMs such as T5 \cite{raffel2023exploring} and BART \cite{lewis-etal-2020-bart}, which have shown success in purely language-based tasks such as summarization and question-answering.


If understanding a model's shortcomings is the first step to ``chipping away at the block,'' addressing those shortcomings is the next natural step. Almost all LLM-based approaches can be thought of as involving a combination of ``data components'' and ``modeling components.''
Data components include the pre-training corpus, the fine-tuning corpus, and any data augmentation strategies used. Modeling components, on the other hand, include the fine-tuning method, prompting strategies, post-processing steps applied over the LLM output, and ensembling methods to combine multiple primitive models.
Most contemporary works on compositional generalization in semantic parsing narrowly focus on one or more of these components (typically, not more than two at a time). For this reason, these works are best interpreted as orthogonal, complementary strategies to improve compositional generalization capabilities.

The paper is structured as follows: Section \ref{sec:definitions} delves into compositional semantic parsing, examining three widely-used benchmark corpora and their compositional generalization definitions. Section \ref{sec:methods} details the literature survey methodology, including search and selection criteria. Section \ref{sec:factors} discusses factors influencing LLM-based semantic parsers' compositional generalization. Section \ref{sec:improve} reviews literature on methods to enhance compositional generalization, covering data-augmentation, neuro-symbolic modeling, and prompt-based methods. Section \ref{sec:discussion} discusses research trends in compositional semantic parsing using LLMs. The paper concludes in Section \ref{sec:conclusion}.

\section{How to Define Compositional Generalization for Semantic Parsing?} \label{sec:definitions}

Broadly speaking, compositionality can be defined as "the ability to generalize to an infinite number of novel combinations from a finite set of symbols." To concretize this rather abstract notion of compositionality, researchers have proposed several benchmark datasets (along with train-test splits) such that the "primitive" symbols in the training set are combined in "unseen" ways in the test set. The devil, of course, lies in the details, and in this case, it lies in the precise notion of the term "unseen."

Since this review's focus is on the modeling side of the equation, a comprehensive exposition of the various types (and subtypes) of compositional generalization defined in literature falls beyond its scope. Nevertheless, we find it informative to briefly summarize the primary motivating factors behind popular benchmark datasets that are commonly used for evaluating compositional generalization.

We shall take a deeper look at three datasets that are widely in use as evaluation benchmarks in compositional generalization research. In doing so, our primary intention is to specify their specific definitions of compositional generalization and outline the motivation(s) behind their development.

    \paragraph{SCAN} The SCAN dataset \cite{lake2018generalization} is based on a synthetic language with a tiny vocabulary of just 13 primitives (e.g., "left," "jump," "opposite," "right"), with the compounds representing action sequences to move around in a 2D environment. In other words, the logical forms corresponding to statements in this language represents the sequence of primitive actions needed to take place in the environment. The SCAN dataset allows  testing models for at least two forms of generalization.
    First, they try out a "length-based" train-test split to test the generalization ability of various sequence models. Specifically, for this generalization scheme, they design their test set to contain utterances that require the generation of output action sequences with length $\geq$ 25, while the training set only contains action sequences with length $<$ 25.
    Second, they create a scheme in which they expect the model to generalize from a primitive (say, $P_i$) in the training split to its composed forms in the test split. More specifically, the training split comprises all other primitives (say, $\{P_1,...,P_{i-1}, P_{i+1},...\}$) along with their corresponding composed forms.
    
    \paragraph{COGS} \citet{kim2020cogs} build on top of the idea of SCAN to include a much wider range of compositional generalization types such that it more closely reflects those that are present in human natural languages. Unlike SCAN, which includes separate splits for each generalization scheme, COGS manages to combine all its proposed generalization schemes into a single, unified train-test split. Broadly speaking, the splits proposed in their work aim to test candidate models' capability to generalize to novel combinations of known primitives in unknown syntactic roles, deeper recursive structures, transformations of verb classes, etc. Together, these generalization schemes are a superset of those demanded by SCAN.
    
    \paragraph{CFQ} \citet{keysers2019measuring} propose a benchmark dataset for compositional semantic parsing based on the Freebase \cite{bollacker2008freebase} knowledge base. Although works such as \citet{lake2018generalization} and \citet{kim2020cogs} provide manually-crafted dataset splits for evaluating compositional generalization, it is not straightforward to compare model evaluation results across two or more different datasets thanks to the unique definitions of compositional generalization they work with. To circumvent this limitation, \citet{keysers2019measuring} propose a systematic method to quantify the ``degree of compositional generalization'' for a given train-test split. Their method, termed Distribution-Based Compositionality Assessment (DBCA), is based on two principles: 1) the distribution of atoms must be as similar as possible between the train and test splits, and 2) the distribution of compounds must be as different as possible. 

Other commonly used benchmark corpora include GeoQuery \cite{zelle1996learning}, Spider \cite{yu-etal-2018-spider}, and Ecommerce-Query \cite{6-yang-seqzero}. However, we will not cover them here in the interest of space.

\section{Survey Methodology} \label{sec:methods}

We use Google Scholar\footnote{\href{https://scholar.google.com}{scholar.google.com}} and Semantic Scholar\footnote{\href{https://semanticscholar.org}{semanticscholar.org}} to search for relevant papers with the following search term:
``(compositional OR compositionality) (AND (llm OR nlp OR natural language OR semantic parsing OR parsing OR dialog OR dialogue))?''
Our primary inclusion criterion is that the work needs to involve LLMs. In addition, we also use two exclusion criteria. Specifically, we exclude a few papers that work on tabular data, and involve multimodal LLMs.

\section{Factors Affecting Compositional Generalization Performance} \label{sec:factors}

This section explores some of the major findings in recent literature regarding the key factors hindering large language model based semantic parsers from being able to generalize compositionally.

\subsection{Inherent Inability of Vanilla Seq2Seq Models to Generalize Structurally}

\citet{3-yao-structural-generalization} present a data-driven analysis that sheds light on the fine-grained underlying causes of the disparity in performance between natural-language generation tasks and symbolic generation tasks. Their first major finding is that vanilla seq2seq models, which lack structural inductive biases built into them, fail to generalize to unseen structures in the output. Even when trained on an augmented dataset that contains samples demonstrating structural recursion, these vanilla models were seen to generalize only up to the maximum recursive depth occurring in the training corpus.


\subsection{Autoregressive Decoding Step as a Bottleneck}
\citet{3-yao-structural-generalization} also present a second finding pertaining to the question of whether the root cause of insufficient compositional generalization lies in the encoder or the decoder (or both). Using the well-established ``probe task methodology''\footnote{The "probe task methodology" evaluates LLMs by testing them on tailored tasks that assess specific skills or knowledge areas. Researchers present models with inputs to elicit responses that reveal certain abilities, such as grammar or logic. Analyzing these responses helps gauge the LLMs' proficiency in these narrow domains.} over BART and T5 models, the authors establish that although the encoder is successful in capturing the structural information to a full extent, the decoding step is lossy in nature and fails to make full use of it.

\subsection{Distributional Mismatch with the Pretraining Corpus}
Since most LLMs are pretrained on corpora that disproportionately contain natural language (and not symbolic forms), the authors hypothesize that reframing semantic parsing as a text generation task (in which the output is forced into a form closer to natural language) might prove to be distributionally closer to its training corpus, thereby offering a fairer task formulation for model evaluation. To test this hypothesis, \citet{3-yao-structural-generalization} create two benchmarks derived from COGS, in which the samples are represented in the form of 1) extractive question-answering problems and 2) free-form question-answering requiring structural generalization respectively. They find that while reframing the model output to natural language does moderately improve performance for the simpler, extractive task, it does not really move the needle by all that much for the second task.

\subsection{Effect of Model Size and ``Mode of Use'' (Fine-Tuned or Zero-Shot)}
\citet{7-qiu-evaluating} present several findings about the relationship between compositional generalizability and 1) the model size and 2) the mode of use (fine-tuned, prompt-tuned, or zero-shot prompted). They find that the compositional generalization performance of fine-tuned LMs does not really increase (and sometimes even degrades) with size. In contrast, although this relative trend is reversed in the case of in-context learning (positive scaling curve), the absolute performance of larger LMs fall short of that of much smaller fine-tuned LMs. Finally, they find that the scaling curve of prompt-tuned LMs is not only positively steeper than that of in-context learning, but also at times outperforms fine-tuned models in absolute terms.

\section{Methods to Improve Compositional Generalization} \label{sec:improve}
While understanding the shortcomings of traditional approaches is the first step to "chipping away" at the problem, developing methods to address them comes next. In general, LLM-based approaches can be thought of as a combination of two activities: data curation and modeling. Data curation involves deciding on the pre-training corpus, curating a fine-tuning corpus, introducing data augmentation strategies, and so on. Modeling involves things such as designing a fine-tuning procedure, developing appropriate prompting strategies, adding post-processing steps to the model output, and coming up with ensembling techniques. Most works on compositional generalization for semantic parsing typically have a narrow focus, and hence, focus on just one or two of the above aspects.

\subsection{Data-Augmentation Based Methods}
The most common data-augmentation method involves augmenting a dataset of (natural language, logical form) pairs in such a way as to include demonstrations of various types of compositional generalization \cite{4-ye-generating-data}. According to this view, compositional generalization is a learnable skill that does not require adding any additional inductive biases to the model. In other words, the hope is that with sufficient examples of compositional generalization, the LLM will continue to learn the skill of compositional generalization (even if not completely solve it). We will look at a few of these works.

\paragraph{General Purpose Frameworks} \citet{2-akyurek-lex-sym} propose a general-purpose framework for compositional data augmentation, which strives to maximize its generality along at least two critical axes. First, the method is domain-agnostic. By projecting raw samples into a relational algebra, they are able to formulate a general-purpose, symbolic data augmentation algorithm. Second, they also strive for generality in terms of the transformations. Rather than focus on a fixed set of transformation types, this ``meta-approach'' infers transformation functions that retain the underlying compositionality structure of the samples (the authors term these transformation functions as ``homomorphisms'').
The homomorphisms are automatically inferred from a set of samples using a brute-force approach.

\paragraph{}

\subsection{Model-Based Methods}
This section examines two classes of methods for generating structured outputs: first, by incorporating inductive biases into models, and second, using novel prompt-based techniques to mimic compositionality.

\subsubsection{Neuro-Symbolic Models}
Neuro-symbolic approaches to compositional generalization aim to combine the strengths of both neural models and ``symbolic,'' grammar-based approaches.
Here, we take the example of \citet{8-shaw-compositional}.
They propose NQG-T5, a ``Neural parsing model and a flexible Quasi-synchronous Grammar induction algorithm,'' which first attempts to parse the input using a grammar-based approach, and in case it fails, falls back to using a vanilla T5 seq2seq model.
There are other works that use hybrid approaches, but they are out of scope of this work.

\subsubsection{Prompt-Based Approaches}
In contemporary prompt-based approaches, Chain-of-Thought prompting \citep{wei2022chain} is seen as a powerful starting point for developing even more complex and nuanced prompting techniques to elicit robust reasoning.

In the context of compositional semantic parsing, a commonly used prompting strategy is to break down the full parsing problem into multiple sub-steps. The ``Parsing as Paraphrasing'' framework is a special case of this paradigm involving two steps, in which the output of the first step is natural language \cite{shin2021constrained}. More concretely, in the first step, the original utterance is paraphrased into a canonical utterance using a model
$$P_{\theta}(c|u).$$
In the second step, the canonical utterance is transformed into the logical form using a grammar $$m = \textrm{Grammar}(c).$$

\paragraph{Decomposition into Steps}
\citet{10-zhou-least-to-most} develop a novel ``Chain-of-Thought'' style prompting technique, which they term "Least-to-Most (LTM) Prompting," to solve semantic parsing over the SCAN benchmark corpus.
The main underlying idea behind LTM Prompting is to decompose a complex problem into simpler subproblems using an LLM, and recursively solve each subproblem (once again, using LLMs).
The authors show that the \texttt{code-davinci-002} model improves its absolute accuracy over the SCAN benchmark from 16\% while using CoT prompting to at least 99\% over all splits of the dataset using just 14 labelled examples while using LTM Prompting.
To contextualize this result, we note that some of the neuro-symbolic methods in literature require around 15,000 samples to achieve similar results. Further, \citet{9-drozdov-compositional} find that while LTM Prompting demonstrates competitive performance over a simple dataset such as SCAN, it falls short in the face of more complex benchmarks such as COGS \cite{kim2020cogs} and CFQ \cite{keysers2019measuring} for a few reasons:
\begin{enumerate}
    \item Sometimes, the subproblems are context-dependent (e.g., dependent on dialogue context), which is not accounted for in the ``vanilla'' LTM method.
    \item Parsing might depend on background information (e.g., domain-specific knowledge) that might not fully fit within the context-window of the LLM.
    \item Finally, the presence of complex, rich natural language might make the initial decomposition step quite challenging even for LLMs.
\end{enumerate}

To address the aforementioned challenges posed by richer natural language variation, \citet{9-drozdov-compositional} further refine the idea of LTM Prompting by introducing a few changes. The most pertinent one is that rather than perform the problem decomposition ``linearly'' in one step, they propose a tree-based approach, whose structure closely resembles standard syntactic parsing in NLP. Independently, \citet{6-yang-seqzero} develop an ensemble-based method that combines a fine-tuned and a zero-shot LLM to generate individual clauses of a ``canonical utterance,'' before using a grammar to parse it into its corresponding logical form. Two key design choices in their model are:
\begin{enumerate}
    \item They feed the output of the predecessor clauses into the LLMs while predicting the current clause. Through experiments, they show that on the GeoQuery and EcommerceQuery datasets, this leads to improved performance. They hypothesize that this is because each clause needs to be modeled using a separate model.
    \item They find that combining the zero-shot model with the fine-tuned model aids overall performance, and explain it via its superior capability to generalize to unseen entities.
\end{enumerate}

\paragraph{Decomposition into Facts}
Rather than breaking down the original question into subquestions, \citet{1-niu-bridging} propose an alternate scheme in which the question is decomposed into a set of \textit{constraints}, each taking the form of a ``simple'' fact-based statement, comprising a subject, a predicate, and an object. The authors' primary motivation behind breaking down the question is to minimize the distributional difference between synthetically-generated questions used for training and the unseen questions that would appear ``in the wild.'' This approach is effective since in general, it is easy to construct datasets with controlled splits synthetically than it is to source naturally-occurring datasets that satisfy the desired splitting criteria or manually rewrite the synthetic utterances.


\section{Discussion} \label{sec:discussion}

We summarized the findings in literature as to the underlying causes for imperfect compositional generalization by large language models.

We can see that recurring trend across almost all ``model-based approaches'' to improving compositional generalizability in LLM-based parsers involves breaking down computation into multiple sequential steps as opposed to doing things in an end-to-end fashion.
 
Stepping back to see the larger picture, as we saw in Section \ref{sec:definitions}, compositional generalization can be subcategorized into several fine-grained definitions, each of which poses a unique set of challenges from a modeling perspective.
Nevertheless, not only do the underlying ``model pathologies'' responsible for imperfect compositional generalization tend to overlap across multiple generalization types, but they also tend to be orthogonal to one another.
One way of interpreting this orthogonality is to view it as a boon for researchers, who can work independently to tackle each pathology in a piecemeal manner.
This view, of course, should not preclude future researchers from posing more foundational research questions questioning the hegemonic status-quo of the seq2seq paradigm as well as that of generative large language models for the general problem of semantic parsing.


\section{Conclusion} \label{sec:conclusion}
Thanks to the popular and commercial success of instruction-tuned large language models such as ChatGPT, these models have captured the public imagination as being all-powerful and general-purpose panaceas for NLP tasks \cite{saphra2023tragedy}.
While there is certainly significant merit to the hype, certain desirable properties such as compositional generalizability continue to elude even the most powerful state-of-the-art models.
By summarizing some of the fundamental issues present in Transformer-based seq2seq models and efforts to move the needle, this survey provides semantic parsing researchers with an overview of the developments in the field, which can then be used as a starting point for further research.

\bibliographystyle{acl_natbib}
\bibliography{custom,supporting,references}

\end{document}